\documentclass[letterpaper, 10pt, conference]{ieeeconf}

\IEEEoverridecommandlockouts
\usepackage{amsmath}

\usepackage[breaklinks=true,bookmarks=false]{hyperref}
\usepackage{caption}
\usepackage{relsize}
\DeclareCaptionFont{smallsmall}{\small\smaller}
\usepackage{graphicx}
\usepackage{float}
\usepackage{siunitx}

\usepackage{cite}
\usepackage{color}
\usepackage{amssymb}
\usepackage{svg}
\usepackage{float}
\usepackage{subcaption}

\makeatletter
\let\NAT@parse\undefined
\makeatother

\usepackage{multirow}
\usepackage{tablefootnote}
\usepackage{threeparttable}
\usepackage{booktabs}
\graphicspath{ {./figures/} }

\usepackage{epstopdf}
\usepackage{makecell}

\usepackage[breaklinks=true,bookmarks=false]{hyperref}

\title{\LARGE \bf
Multimodal Indoor Localization Using Crowdsourced Radio Maps
}

\author{Zhaoguang Yi\authorrefmark{4}\authorrefmark{2}, Xiangyu Wen\authorrefmark{4}\authorrefmark{2}, Qiyue Xia\authorrefmark{4}\authorrefmark{2}, Peize Li\authorrefmark{4}\authorrefmark{2}, Francisco Zampella\authorrefmark{5}, \\Firas Alsehly\authorrefmark{5}, Chris Xiaoxuan Lu\authorrefmark{1}\authorrefmark{3}
\thanks{\authorrefmark{1}Corresponding author. Email: {\tt\footnotesize xiaoxuan.lu@ucl.ac.uk}}
\thanks{\authorrefmark{4}Denotes equal contribution.}
\thanks{\authorrefmark{2}School of Informatics, University of Edinburgh, United Kingdom.}
\thanks{\authorrefmark{5} Huawei Technologies (UK).}
\thanks{\authorrefmark{3} Department of Computer Science, University College London.}
}

\begin{document}
\maketitle

\pagestyle{empty}

\begin{abstract}
Indoor Positioning Systems (IPS) traditionally rely on odometry and building infrastructures like WiFi, often supplemented by building floor plans for increased accuracy. However, the limitation of floor plans in terms of availability and timeliness of updates challenges their wide applicability. In contrast, the proliferation of smartphones and WiFi-enabled robots has made crowdsourced radio maps – databases pairing locations with their corresponding Received Signal Strengths (RSS) – increasingly accessible. These radio maps not only provide WiFi fingerprint-location pairs but encode movement regularities akin to the constraints imposed by floor plans. This work investigates the possibility of leveraging these radio maps as a substitute for floor plans in multimodal IPS. We introduce a new framework to address the challenges of radio map inaccuracies and sparse coverage. Our proposed system integrates an uncertainty-aware neural network model for WiFi localization and a bespoken Bayesian fusion technique for optimal fusion. Extensive evaluations on multiple real-world sites indicate a significant performance enhancement, with results showing $\sim25\%$ improvement over the best baseline.
\end{abstract}

\section{Introduction}
Indoor location awareness is a key capability for smart devices and a fundamental need in embodied intelligence. There is a surge in applications of Indoor Positioning Systems (IPS) ranging from smart retail navigation to space utilization analysis. 

Due to the non-line-of-sight issue of GPS in indoor environments, existing IPS leverage inertial measurement units (IMU) and/or building-dependent infrastructures such as WiFi, Bluetooth or magnetic fingerprints to localize smart devices. More recently, building floor plans emerge to be used as an effective input modality for IPS and can be fused with IMU and other building-dependent infrastructures to improve the localization accuracy \cite{herath2021fusiondhl}\cite{xiao2014lightweight}. Intuitively, floor plans constrain the
movement path of a moving agent - people cannot walk through walls and can only enter a room through a door, and can serve as strong cues in correcting localization offsets. However, the floor plans of many environments can be absent or lack timely updates after map creation. As a result, the usefulness of such indoor localization methods is largely limited by the availability of accurate maps.

\begin{figure}[!t]
\centering
\captionsetup{belowskip=-20pt}
\includegraphics[width=1.0\linewidth]{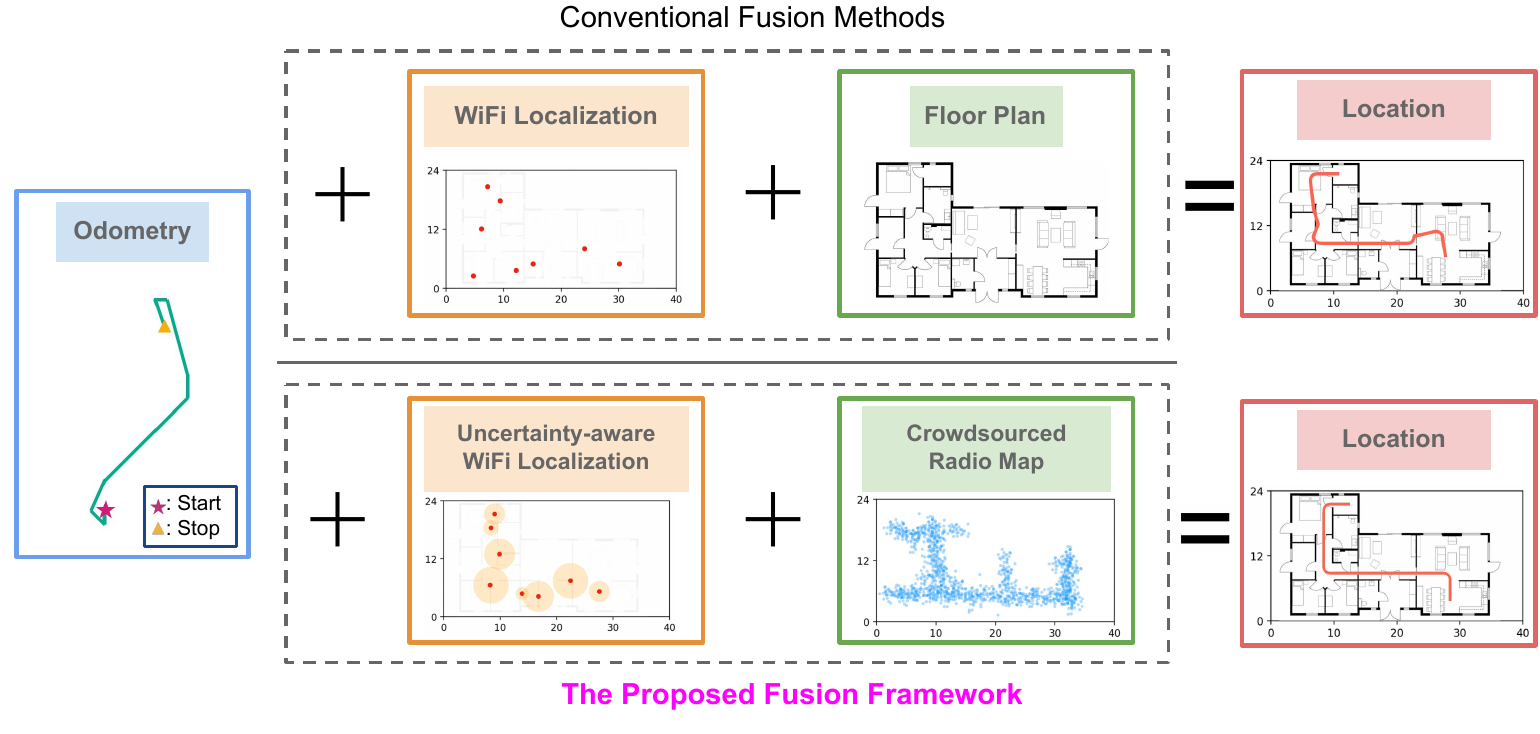}
\caption{Conventional IPS uses accurate radio maps for WiFi localization and floor plans in fusion. To fully utilize the noisy crowdsourced radio map, we propose an uncertainty-aware WiFi localization module and a bespoken Bayesian fusion method that replaces floor plans with crowdsourced radio maps in fusion. }
\label{fig:openfig}
\end{figure}

On the other side, with the proliferation of smartphones and WiFi-enabled mobile robots, the radio maps of WiFi become largely available from crowdsourcing. A radio map broadly refers to a database 
consisting of a number of paired locations and their respective Received Signal Strengths (RSS) from different wireless access points. A research question of our interest is - \emph{can we replace floor plans with radio maps in multimodal IPS?} 
Indeed, crowdsourced radio maps are often derived from tons of walking/moving trajectories by various users. Therefore, they not only contain wireless and location pairs but also carry side information about traversing regularity by inspecting the locations of fingerprints. With sufficient crowdsourced trajectories, we hypothesize the regularity embedded in a radio map implicitly reveals the allowable paths and can be used as an effective cue for localization, sharing a similar spirit of floor plans in constraining localization results.
However, due to the nature of crowdsourcing, radio maps created this way inevitably suffer from location inaccuracy and sparse position coverage. Prior arts in fingerprint-based WiFi IPS \cite{huang2019online} found that the accuracy of the WiFi localization systems can be far from satisfying and impose challenges in fusion if built upon noisy crowdsourced radio maps. Meanwhile, the sparse position coverage of radio maps cannot provide location constraints as explicit as that of floor plans, limiting the adoption of map-matching-like fusion methods \cite{xiao2014lightweight,herath2021fusiondhl}.  

In this work, we propose a new framework to deal with the issues of radio map issues and unravel their full potential in IPS. Our framework, illustrated in Fig.~\ref{fig:openfig}, includes an uncertainty-aware neural network model for WiFi fingerprinting with radio maps and a new Bayesian filtering method admissible to the fusion of radio maps. Qualitative and quantitative evaluations on three real-world sites demonstrate that the proposed system is able to
produce $\sim25\%$ performance gain compared to the state-of-the-art IPS.

\section{Related Work}

\noindent \textbf{Indoor Localization via Wireless Networks}
Indoor localization based on existing wireless communication networks, such as WiFi \cite{wifireview2015yang} and Bluetooth\cite{bluetoo2016chen}, based on their RSS, has gained popularity because it does not require additional hardware \cite{wifireview2015yang}. For such systems, two primary methodologies are employed:
1) Utilizing an access-point database that comprises access-point IDs and their respective geo-coordinates. This setup facilitates the determination of user location through trilateration or weighted averaging of access-point coordinates \cite{WiGLEonline}.
2) Employing a fingerprint database, a radiomap, which records the received signal strength indicators and the corresponding geo-coordinates at their observation points\cite{radar2000bah},\cite{WILL2013wu}. Such an approach stands out for its inherent scalability, amplified by the prospects of crowd-sourced data collection \cite{crowdsource2010mink}. 


\noindent \textbf{Uncertainty Estimation}
Estimating the uncertainty, or more specifically, the suspicious evidence cues derived from data is crucial for the integration in real-world applications and as input for data fusion processes. 
To quantify uncertainty within deep learning, two primary distinctions emerge: epistemic uncertainty, associated with the model, and aleatoric uncertainty, related to the data \cite{kendall2017uncertainties}. 
Estimating aleatoric uncertainty can enhance the robustness of the network when training on noisy datasets. 
By introducing an additional aleatoric uncertainty prediction head, the reliability of neural networks has been significantly strengthened across various applications \cite{cai2022stun}\cite{poggi2020uncertainty}\cite{taha2019unsupervised}\cite{feng2018safe}. Our research primarily concentrates on enhancing localization performance on noisy radio maps by predicting uncertainty, which in turn aids in data fusion processes.

\noindent \textbf{Multi-modal Sensor Fusion}
Indoor positioning and navigation techniques using smartphones have seen significant advances with the incorporation of various data sources and modeling methods. WiFi-SLAM \cite{ferris2007wifi,mirowski2013signalslam} relays on WiFi-based loop-closing constraints, which effectively minimize the accumulation errors commonly found in inertial navigation. Bayesian filter-based algorithms can also be used to perform the multi-modal sensor fusion to estimate the online position, such as particle filter\cite{zou2017Accurate,Hong2014WaP,carrera2016real}, Kalman filter and its variant\cite{Hellmers2013IMU,Poulose2019Indoor,Feng2020Kalman}. Several studies have incorporated floor plan information into their fusion systems via offline image processing techniques \cite{herath2021fusiondhl,ma2017pedestrian}, while other methods use the floor plan as a constraint for particle filtering \cite{Hong2014WaP,carrera2016real} and conditional random fields\cite{xiao2014lightweight}. However, there is no method to properly fuse IMU, uncertainty-aware WiFi localization results, and the crowdsourced WiFi radio map data.

\section{Problem Formulation}
Presume we have an offline crowdsourced radio map containing $N$ fingerprints and their associated locations $\mathcal{M} = \{(\mathbf{G}_i, \mathbf{L}_i)|i=1,2,..,N\}$, where $\mathbf{L}_i$ denotes the location coordinates and $\mathbf{G}_i = \{(MAC_j, RSS_j)|j=1,2,..,|\mathbf{G}_i|\}$ is called a WiFi fingerprint represented as a list of MAC addresses of the WiFi Access Points (APs) and their corresponding RSS values at location $\mathbf{L}_i$. Note that the number of WiFi APs $|\mathbf{G}_i|$ varies across locations and the locations $\mathbf{L}_i$ considered in this work are two-dimensional, namely $\mathbf{L}_i= (x_i, y_i)$. As the radio map is crowdsourced rather than manually surveyed, the association between WiFi fingerprints and location coordinates can be inaccurate, rendering a \textbf{noisy radio map}.
During the online stage and timestamp $t$, a mobile agent at an unknown location will measure a fresh WiFi fingerprint measurement $\mathbf{G}_t$ and have a location/pose change $\Delta\mathbf{\check{L}}_{t-1:t}$. predicted from an odometry module. Our objective here is to obtain an updated location coordinate $\mathbf{\hat{L}}_t$ by utilizing the noisy crowdsourced map $\mathcal{M}$, the fresh fingerprint measurement $\mathbf{G}_t$ and the location change $\Delta\mathbf{\check{L}}_{t-1:t}$. Without losing the generality, the mobile agent considered in our task can be either a smartphone or a WiFi-enabled mobile robot. In the case of smartphones, the predicted location or pose change can come from the smartphone inertial sensor and off-the-shelf inertial odometry e.g., \cite{imupdr2013Li};
In the case of mobile robots, such prediction change can be accessed from their wheel odometry module.

We propose a new method to address the problem above. At a high level, our method consists of an uncertainty-aware WiFi localization neural network model (c.f. Sec.~\ref{sec:uncertainty_aware}) and a bespoken Bayesian fusion approach (c.f. Sec.~\ref{sec:multimodal_fusion}) that combines the WiFi localization results, odometry estimates and the noisy radio map to derive a more robust location.

\begin{figure*}[!t]
    \centering  
    \captionsetup{belowskip=-10pt}
    \includegraphics[width=0.85\linewidth]{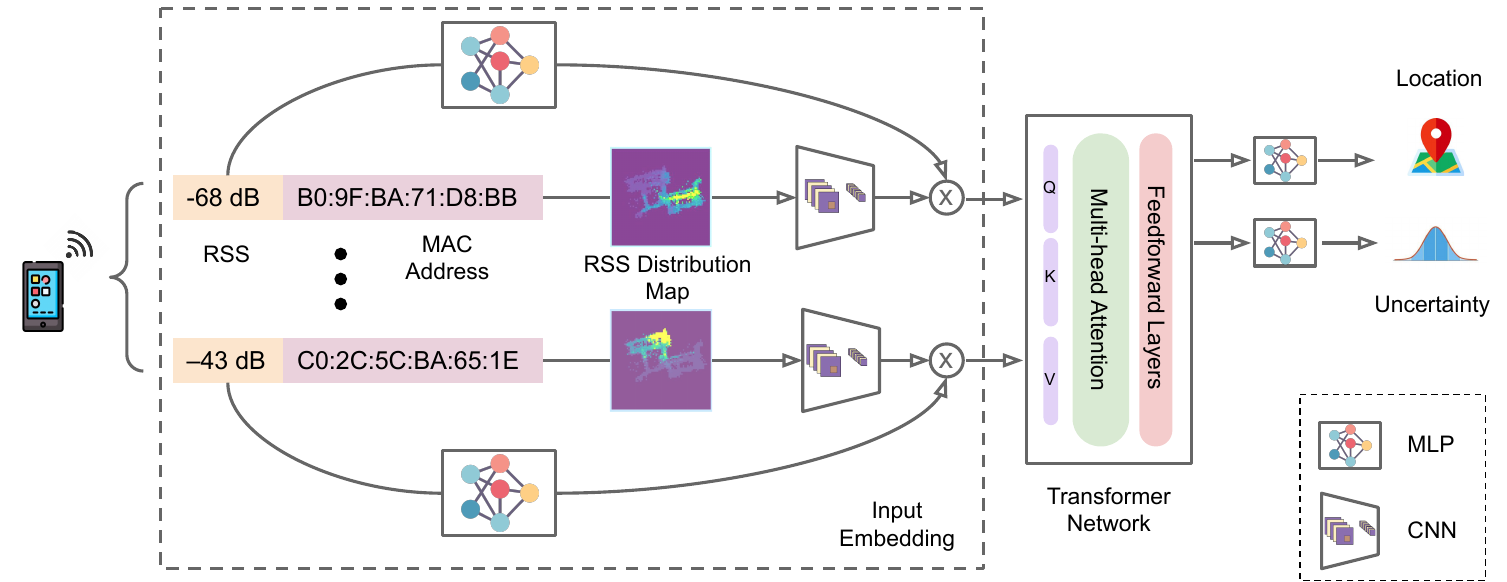}
    \caption{The proposed \emph{end-to-end trainable} WiFi localization model. Given a WiFi fingerprint list measured by a WiFi-enabled device (the smartphone in the figure), the MAC address part is input-embedded by applying a CNN on its AP-associated RSS distribution map; The RSS part of the fingerprint is embedded via an MLP, and multiplied with the MAC address embedding (c.f. Sec.~\ref{sub:input_embedding}). The constituted fingerprint embeddings are then fed into the transformer featuring a multi-head attention block in it (c.f. \ref{sub:multi_head_attention}). After obtaining the encoded fingerprint features by the transformer, we append two MLPs to regress the location coordinates and quantify the prediction uncertainty (c.f. Sec.~\ref{sub:uncertainty_estimation}) for subsequent fusion.}
    \label{fig:wifi_overview}
\end{figure*}
\section{Uncertainty-aware WiFi Localization}
\label{sec:uncertainty_aware}

\subsection{Design Motivation}
We start with the motivation behind our WiFi localization model. As found in \cite{wifioverview2017xia} 
, the lengths of WiFi fingerprints $\mathbf{G}_i$ vary across locations and time even at the same location. This requires the WiFi localization model to take WiFi fingerprints in different lengths but still be able to output location coordinates. This requirement is similar to the one of natural language processing, where language models often take input as the sentences in different token lengths, but are still able to perform inference. Inspired by recent successes of transformers in natural language processing \cite{vaswani2017attention,devlin2018bert}, 
we propose utilizing the transformer's encoder in our network design to handle the varying-length WiFi fingerprints.
Moreover, to deal with the inherent noisy match in the crowdsourced radio map, we further introduce a network block to model the prediction uncertainty.


\subsection{Input Embedding of WiFi Fingerprint}
\label{sub:input_embedding}
A fingerprint in the list format of $(MAC_j, RSS_j)$ cannot be directly employed as the input to a transformer-like network. 
Akin to the idea of distributed representations of words in natural language processing \cite{mikolov2013efficient}, 
a fingerprint needs to be represented as a continuous vector before being fed into the model. 
Pre-trained word embedding layers are often available for language models by which the derived word embeddings are close to each other if two words have similar semantic meanings. 
However, such a semantic-preserving embedding layer is absent in the case of WiFi fingerprints esp. for the MAC address of APs. The challenge lies in that it is difficult to relate two nearby APs by their MAC address manifestation - which is simply a 12-character alphanumeric attribute that is used to identify individual electronic devices. 

To inject the location vicinity semantics in MAC address embedding, we propose to utilize the RSS distribution map.
For each MAC address, we retrieve the radio map and construct a heatmap using the RSS signal strength received from that MAC address at all available locations. Intuitively, the highlighted area approximately indicates the location of the MAC address, and nearby WiFi APs tend to have similar RSS distribution maps. CNN blocks are used to process the distribution maps and extract the dense vectors as MAC address embeddings. 
As we will see in Sec.\ref{subs:input_mac_embedding} the physical vicinity can be conveyed in the similarity of the embeddings. Even for the same AP, its RSS values measured by a mobile agent can vary across places. To make use of the RSS value in the WiFi fingerprint, we introduce a learnable MLP module that maps the RSS value and multiplies with the MAC address embedding to derive the final fingerprint embedding, $\mathbf{E}_i$. Fig.~\ref{fig:wifi_overview} illustrates the input embedding. 
\subsection{Multi-Head Attention Block}
\label{sub:multi_head_attention}
We aim to accurately predict location coordinates based on WiFi fingerprint embeddings. Given the dynamic and complex nature of WiFi fingerprints, it's crucial to capture dependencies between different features of the fingerprint for a precise location prediction. 
We thus adopt the self-attention transformer. Essentially, it allows each feature in the WiFi fingerprint embedding to weigh the importance of other features based on their content. This is particularly advantageous for our task because certain WiFi fingerprints might carry more relevance in determining specific location coordinates than others. Diving deeper into the mechanics, let's first revisit the single-head attention mechanism:
\begin{equation}
\begin{aligned}
head_h = Attention(\mathbf{E}_i \mathbf{W}_{Q_h}, \mathbf{E}_i \mathbf{W}_{K_h}, \mathbf{E}_i \mathbf{W}_{V_h}) \\
= softmax\left(\frac{\mathbf{E}_i \mathbf{W}_{Q_h}(\mathbf{E}_i \mathbf{W}_{K_h})^T}{\sqrt{d_k}}\right)\mathbf{E}_i \mathbf{W}_{V_h}
\end{aligned}
\end{equation}
where $\mathbf{E}_i$ is the input MAC embedding; $\mathbf{W}_{Qi}$, $\mathbf{W}_{Ki}$, and $\mathbf{W}_{Vi}$ are the projection matrices for query, key, and value respectively; $\sqrt{d_k}$ is the dimension of the key $k$ and query $q$.
However, relying on a single attention head might restrict the model from capturing multiple types of relationships or dependencies between the features. This is where the multi-head attention mechanism, as proposed in \cite{vaswani2017attention} is employed:
\begin{equation}
\begin{aligned}
MultiHead(\mathbf{E}_i \mathbf{W}_{Q_h}, \mathbf{E}_i \mathbf{W}_{K_h}, \mathbf{E}_i \mathbf{W}_{V_h})\\=Concat(head_1, \ldots, head_H) \cdot \mathbf{W}_O
\end{aligned}
\end{equation}
where $head_1, \ldots, head_H$ represent the $H$ single-head attention results. By using multiple attention heads, the model can simultaneously attend to various subspace representations of the input. Each head potentially captures different types of dependencies within the fingerprint embeddings. 
These concatenated results are then combined using a linear transformation matrix $\mathbf{W}_O$, to produce an attended embedding. Each attended embedding is applied a non-linear transformation before output. After processing through several of such multi-head attention layers, the output from the last layer is further refined by a feed-forward network to regress the location coordinates.

\subsection{Uncertainty Estimation} 
\label{sub:uncertainty_estimation}
As mentioned earlier, crowdsourced radio maps often come with inherent noise. Using them directly to train a transformer localization model can lead to high susceptibility to errors in predictions, given the model might learn from the noise rather than the actual signal. As a natural response to this challenge, one might consider ways to make the model more resilient to noise and simultaneously offer a more informative prediction. Thus, instead of merely giving a point estimate that provides a singular location prediction, it would be more beneficial if our model could quantify the level of confidence in its predictions. 

To this end, we reformulate the network's output of location coordinates as a \emph{probability distribution}. We choose the Gaussian distribution for this purpose, given its well-understood properties and ease of integration in subsequent fusion steps:
\begin{equation}
\mathbf{L}^{G}_i \sim \mathcal{N}(f_{\mu}(\mathbf{G}_i), f_{\sigma}(\mathbf{G}_i))
\end{equation}
where $f_{\sigma}(\cdot)$ is an extra neural network head capturing the variance which is pivotal as it provides an estimate of the uncertainty associated with the prediction. By transitioning to this probabilistic framework, we ensure the transformer model does not blindly commit to a potentially noisy sample. Instead, it offers a range of possibilities, weighted by their likelihood. This not only makes the model less prone to overfitting to the noise but also enriches the output, as the variance gives an insight into the reliability of the prediction. 

Altogether, given crowdsourced a radio map $\mathcal{M} = \{(\mathbf{G}_i, \mathbf{L}_i)|i=1,2,..,N\}$, we train our WiFi localization neural network with an uncertainty-aware loss function:
\begin{equation}
\mathcal{L} = \sum_{i}^{N} \frac{\lVert f_{\mu}(\mathbf{G}_i) - \mathbf{L}_i \rVert_2}{f_{\sigma}(\mathbf{G}_i)}+ln{f_{\sigma}(\mathbf{G}_i)} 
\end{equation}

During training, the first term drives predictions closer to the radio map location coordinates. The numerator of the first term is to ensure that a larger variance is given for samples with significant prediction errors, mitigating large gradient impacts. The second term acts as a regularizer, discouraging the network from consistently yielding high variance.
\section{Multimodal Fusion}
\label{sec:multimodal_fusion}

\subsection{Design Motivation}
We begin by presenting the motivation behind our multimodal fusion method which combines the strengths of the extended Kalman filter and particle filters. Prior arts of particle filters have shown the effectiveness of using particle filters to combine the WiFi localization results, inertial odometry and floor plans. In our context, however, the floor plan is unavailable and we can only access the radio map via crowdsourcing. A straightforward idea is to use the kernel density estimation algorithm\cite{silverman2018density} to generate a continuous likelihood distribution over the map area 
  , indicating the frequency and accessibility of specific locations. This information can act as a constant similar to floor plans by adjusting the weight of importance in particle filters. However, since radio maps provide only a sparse estimate of location frequency, the resulting distribution is uneven. 
Consequently, without particles in true but unknown areas, the \emph{kidnapped robot problem} \cite{Thrun2005Probabilistic,hester2008negative} is likely to happen if we are confined to changing the importance weight only (see the supplementary video). To address this, we propose incorporating the Kalman correction step into particle filters, creating an Extended Kalman Particle Filter (EKPF) that includes model prediction, importance weight adjustment, and WiFi measurement correction. This approach aims to ensure particles are more aligned with WiFi localization findings.

\subsection{Model Prediction} 
In EKPF, every particle maintains states of the particle $\mathbf{s}_{t}=[x_t,y_t,\theta_t,v_t]^T$, an importance weight $w$ and a corresponding error covariance matrix $\mathbf{P}_{t}$, where $x_t,y_t$ are the map frame 2D position, $\theta$ represents the odometry frame to the map frame rotation. $v$ is the scale ratio of the estimated velocity from IMU to the filter velocity. Whenever relative position updates $\mathbf{O_{t}} = {[O_{tx},O_{ty}]^T}$ in odometry frame is received, the states are propagated as follows:
\begin{equation}
\begin{aligned}
\mathbf{s}_{t}=\mathbf{s}_{t-1}+\mathbf{M}\cdot(\mathbf{O_{t}} \cdot (1+\phi(n_p)))+ [0,0,\phi(n_r),0]^T 
\end{aligned}
\end{equation}
\begin{equation}
\begin{aligned}
\mathbf{M} = {\left[\begin{array}{l l l}{v_t\cos\theta_{t-1}}&{-v_t\sin\theta_{t-1}}\\ {v_t\sin\theta_{t-1}}&{v_t\cos\theta_{t-1}}\\ {0}&{0}\\ {0}&{0}\end{array}\right]}
\end{aligned}
\end{equation}
where $\phi(n)$ represents random Gaussian noise with zero mean and variance $n$.

\begin{equation}
\begin{aligned}
\check{\mathbf{P}}_{t}=\mathbf{F}_{t-1}\check{\mathbf{P}}_{t-1}\mathbf{F}_{t-1}^{T}+\mathbf{N}_{t-1}\mathbf{Q}_{t-1}\mathbf{N}_{t-1}^{T}
\end{aligned}
\end{equation}
where $\check{\mathbf{P}}_{t}$ is the state covariance at time $t$, $Q_{t-1}$ is the noise covariance at time $t$. ${F}_{t-1}$ is the motion model Jacobian with respect to the last state while ${N}_{t-1}$ is the motion model Jacobian with respect to the noise of the last state. 

\subsection{Importance Weight Update} 
\label{sub:importance}
The weight of every particle is updated using its current location and the
likelihood map during this stage. To obtain a smooth prior map $f_{map}$, the crowdsourced radio map is fed into the Gaussian kernel density estimation algorithm \cite{silverman2018density} and then normalized:
\begin{equation}
f_{map}(\mathbf{L}) = \frac{1}{N} \sum_{i=1}^{N} \frac{1}{\sqrt{2\pi}} e^{-\frac{1}{2} \left( {L - L_i} \right)^2} + \beta
\end{equation}
where $\beta$ is a small constant to control the smoothness of the generated prior map.
The importance weight of every particle is updated using its current location states.
Like standard particle filter algorithms, the ratio of effective particles is monitored during the process and the standard systematic method \cite{doucet2001sequential} is used for the particle resampling. Note that during the particle resampling phase, the error covariance matrix \( \mathbf{P} \) remains unchanged.




\subsection{Measurement Model}

Similar to the usage in the original Kalman filter, in EKPF, the measurement model relates the current state $\mathbf{\check{s}}_t$ to a fresh WiFi localization result at timestamp $t$, $\mathbf{L}^{G}_t \sim \mathcal{N}(f_{\mu}(\mathbf{G}_t), f_{\sigma}(\mathbf{G}_t))$.
In the map frame, we have:
\begin{equation}
\begin{aligned}
 f_{\mu}(\mathbf{G}_t)=\left[\begin{array}{c c c c}{{1}}&{{0}}&{{0}}&{{0}}\\ {{0}}&{{1}}&{{0}}&{{0}}\end{array}\right]\mathrm{s}_{t}+{\bf m}_{t}\;
\end{aligned}
\end{equation}
$\bf{m}_{t}$ is the measurement noise.
To compute the Kalman Gain, we have:
\begin{equation}
\begin{aligned}
\mathbf{K}_t = \mathbf{\check{P}}_t \mathbf{H}_t^T \left(\mathbf{H}_t \mathbf{\check{P}}_t \mathbf{H}_t^T +  \mathbf{R}_t  \right)^{-1} 
\end{aligned}
\end{equation}
where $R_t$ is the measurement noise covariance where we used the estimated uncertainty $f_{\sigma}(\mathbf{G}_t)$ from the aforementioned WiFi localization model; $\mathbf{H}_t$ is the measurement model Jacobian with respect to the last state. The Kalman Gain $K_t$ is used to correct the predicted state:
\begin{equation}
\begin{aligned}
\mathbf{\hat{s}}_t = \mathbf{\check{s}}_t + \mathbf{K}_t \left(\mathbf{L}_t -  f_{\mu}(\mathbf{G}_t)\right)
\end{aligned}
\end{equation}
Finally, the covariance correction is updated as follows:
\begin{equation}
\begin{aligned}
\mathbf{\hat{P}}_t = \left(\mathbf{I} - \mathbf{K}_t \mathbf{H}_t \right)\mathbf{\check{P}}_t
\end{aligned}
\end{equation}
where $\mathbf{I}$ represents the identity matrix.

\section{Experimental Setup}

\subsection{Data Collection}

In diverse campus environments, we gathered WiFi fingerprints and IMU data using multiple smartphones over varying durations. This data was refined using a graph optimization algorithm \cite{huaweiradiomap} to produce radio maps for different structures. We focused on Building A (2,800 $m^2$ commercial space), Building B (3,300 $m^2$ educational facility), and Building C (4,500 $m^2$ library). Ground truth was acquired using Velodyne VLP-16 LiDAR integrated with an Xsens IMU, offering precise location data at 200Hz and we ran with Lio-SAM \cite{liosam2020shan} to obtain the true pose. The smartphones collected IMU data at 100Hz and WiFi fingerprints every 2-3 seconds. We gathered around 2 hours of test data across the buildings, with trajectories lasting 5-10 minutes and covering a total of $4.2$ km.


\subsection{Implementation Details}

\subsubsection{WiFi Localization Network}
The model was end-to-end trained on the radio map dataset, split 9:1 for training and validation. The best model is selected based on the performance achieved on the validation set. Note that the test set for performance evaluation is separate and independent of the radio map set. Coordinates were min-max normalized for faster convergence. Training used the AdamW optimizer \cite{adamw2019decoupled} at a $1e^{-4}$ learning rate, betas of $(0.9, 0.999)$, and a weight decay of $1e^{-3}$ over 200 epochs. 
Unless otherwise specified, the ReLU\cite{relu2019deep} activation function was utilized throughout the model. The CNN used for MAC address embedding consisted of two layers with kernel sizes 20 and 5 and strides 4 and 2, raising channel counts from 1 to 16. An MLP then reduced dimensions to $256*1$.
The RSS embedding MLP has neurons arranged as $(1, 128, 64, 1)$, using Sigmoid activation at the output. The multi-head attention transformer has 4 layers and 4 heads, a feed-forward size of 256, and a $0.2$ dropout rate. Both output layers's hidden dimensions are set as $(256, 128, 64)$. The uncertainty-focused MLP's output adopts the Sigmoid function mapping the probability from $[0, 1]$. The localization MLP's last layer only has the linear transformation.

\subsubsection{EKPF Fusion}

$400$ particles were used in all experiments. Empirically, the location noise covariance, $\phi(n_p)$, was set to $0.1$, and the rotation covariance, $\phi(n_r)$, was set to $0.05$. Two critical EKPF parameters are: the softening constant $\beta$ for the prior map likelihood and the measurement noise covariance constant $\gamma$ adjusted by $f_{\sigma}(\mathbf{G}_t)$. Given the inconsistent reliability of the prior map and WiFi model across buildings, optimal parameters differ. For each site, a validation trajectory was used for a grid search to determine these settings, with $\beta$ ranging from $1e^{-4}$ to $1e^{-5}$ and $\gamma$ from 100 to 200. The kernel density function processed the radio map using a Gaussian kernel with a bandwidth of $1.0$.

\section{Results}

\subsection{Multimodal IPS Performance}

We now compare our multimodal localization approach with the following competing methods: (1) \textbf{FusionDHL}\cite{herath2021fusiondhl}: We adapt the FusionDHL to a version without floorplan on our dataset. FusionDHL is an offline optimization algorithm based on least squares and the closest baseline to ours. (2) \textbf{EKF}\cite{kalman1960new}: The normal extended Kalman filter combines the odometry and the uncertainty-aware WiFi localization results. (3) \textbf{PF}\cite{doucet2001sequential}: The particle filter fusion method uses an implementation similar to \cite{zou2017Accurate}, fusing the odometry and the uncertainty-aware WiFi localization results.

\begin{figure*}[t]
     \centering
     \begin{subfigure}[b]{0.325\textwidth}
         \centering
         \includegraphics[width=\textwidth]{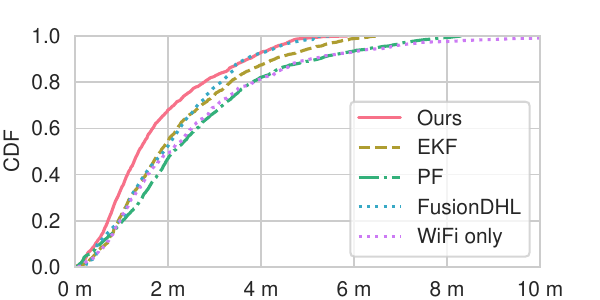}  
         \caption{Building A. Mean error of ours: 1.71m}
         \label{fig:y equals x}
     \end{subfigure}
     \begin{subfigure}[b]{0.325\textwidth}
         \centering
         \includegraphics[width=\textwidth]{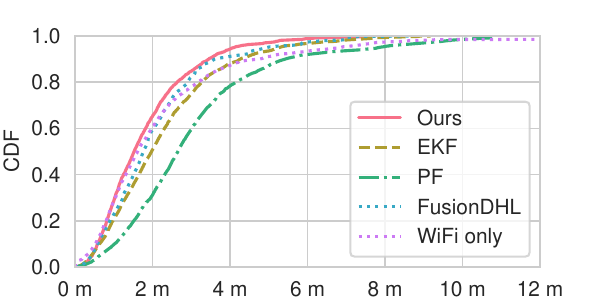}
         \caption{Building B. Mean error of ours: 1.91m}
         \label{fig:three sin x}
     \end{subfigure}
     \begin{subfigure}[b]{0.325\textwidth}
         \centering
         \includegraphics[width=\textwidth]{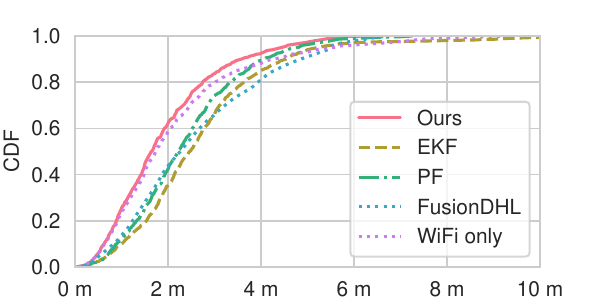}
         \caption{Building C. Mean error of ours: 1.82m}
         \label{fig:five over x}
     \end{subfigure}
               \captionsetup{belowskip=-10pt}
        \caption{Overall performance of our multimodal IPS.}
        \label{fig:cdf}
        \vspace{-0.007\textwidth}
\end{figure*}

\begin{figure*}
     \centering
     \begin{subfigure}[b]{0.325\textwidth}
         \centering
         \includegraphics[width=0.9\textwidth]{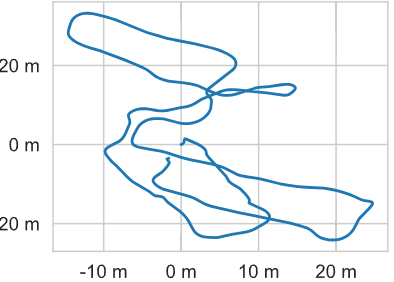}
         \caption{Raw Odometry}
         \label{fig:y equals x}
     \end{subfigure}
     \begin{subfigure}[b]{0.325\textwidth}
         \centering
         \includegraphics[width=0.9\textwidth]{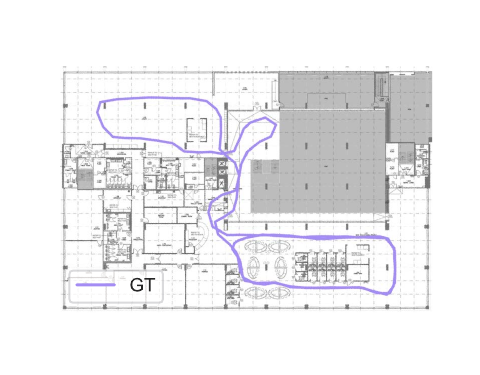}
         \caption{Ground Truth}
         \label{fig:three sin x}
     \end{subfigure}
     \begin{subfigure}[b]{0.325\textwidth}
         \centering
         \includegraphics[width=0.9\textwidth]{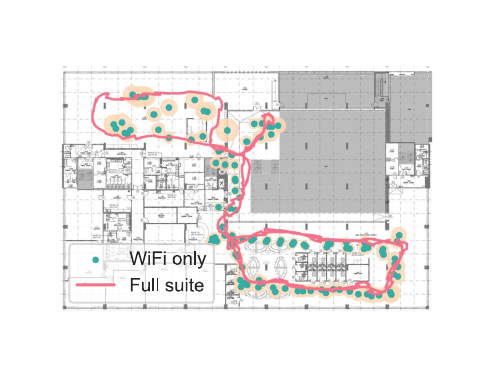}
         \caption{Our method (ring radius corresponds to uncertainty)}
         \label{fig:five over x}
     \end{subfigure}
        \caption{Qualitative results of our method in the library (Building C). }
        \label{fig:result_example}
        
\end{figure*}

Figure~\ref{fig:cdf} illustrates the cumulative distribution function for localization errors in meters, excluding data from the first minute to allow filter methods time to stabilize. Our Extended Kalman Particle Filter (EKPF) fusion approach outperforms all baseline methods across three sites. Compared to FusionDHL, EKPF better integrates radio map data with WiFi/IMU fusion, enhancing localization accuracy by approximately 10.6\%, 21.5\%, and 43.3\% in three different buildings. Figure~\ref{fig:result_example} further highlights the qualitative improvements our method achieves in the largest building.

\subsection{WiFi Localization Performance}

We evaluate our model against both traditional and state-of-the-art machine-learning localization techniques. For the conventional method, we used the Weighted K-nearest neighbours (WKNN)\cite{wknn2004he}, which employs Euclidean distance with \(k=50\) neighbours. Additionally, we compared against a method utilizing a stacked denoising autoencoder for regression (DSAE)\cite{DSAE2019wa}, which first trains an autoencoder and then fine-tunes with an MLP for improved localization.

In Tab.~\ref{tab:wifi-results}, our model's performance outshines both WKNN and DSAE across three buildings, excelling in mean, median, and 95th percentile errors. Although we observed a slightly higher maximum error in Building B compared to WKNN, our model's uncertainty prediction effectively reduces outlier impact, ensuring dependable outcomes overall.

\begin{table}[!h]
    \centering
    \caption{WiFi Localization Performance Comparison}
    \label{tab:wifi-results}
    \begin{threeparttable}
    \begin{tabular}{llllll} 
    \hline
    Site & Method & Mean & Median & Max & 95th \\ 
    \hline
    \multirow{3}{*}{Building A} & WKNN\cite{wknn2004he}& 6.0200 & 5.0613 & 40.5234 & 14.1351 \\
    & DSAE\cite{DSAE2019wa}& 5.2531 & 4.4546 & 19.8080 & 11.8786 \\
    & Ours & \textbf{2.5706} & \textbf{2.1238} & \textbf{14.4922} & \textbf{6.3661} \\ 
    \hline
    \multirow{3}{*}{Building B} & WKNN\cite{wknn2004he}& 4.3299 & 3.4623 & \textbf{19.5425} & 10.3376 \\
    & DSAE\cite{DSAE2019wa}& 5.5253 & 5.1184 & 22.1047 & 11.5780 \\
    & Ours & \textbf{3.3942} & \textbf{2.4951} & 25.3909 & \textbf{9.1316} \\ 
    \hline
    \multirow{3}{*}{Buidling C} & WKNN \cite{wknn2004he} & 7.8455 & 6.4902 & 24.4383 & 18.9681 \\
    & DSAE \cite{DSAE2019wa} & 2.7263 & 2.3757 & 21.7769 & 5.7564 \\
    & Ours &\textbf{2.1596} & \textbf{1.7336} & \textbf{9.4093} & \textbf{5.5427} \\
    \hline
    \end{tabular}
    \end{threeparttable}
    
\end{table}

\subsection{Ablation Study}

\subsubsection{Input MAC Embedding}
\label{subs:input_mac_embedding}

\begin{figure}
    \centering
    \begin{subfigure}{0.23\textwidth}
    \centering
        \includegraphics[width=.8\textwidth]{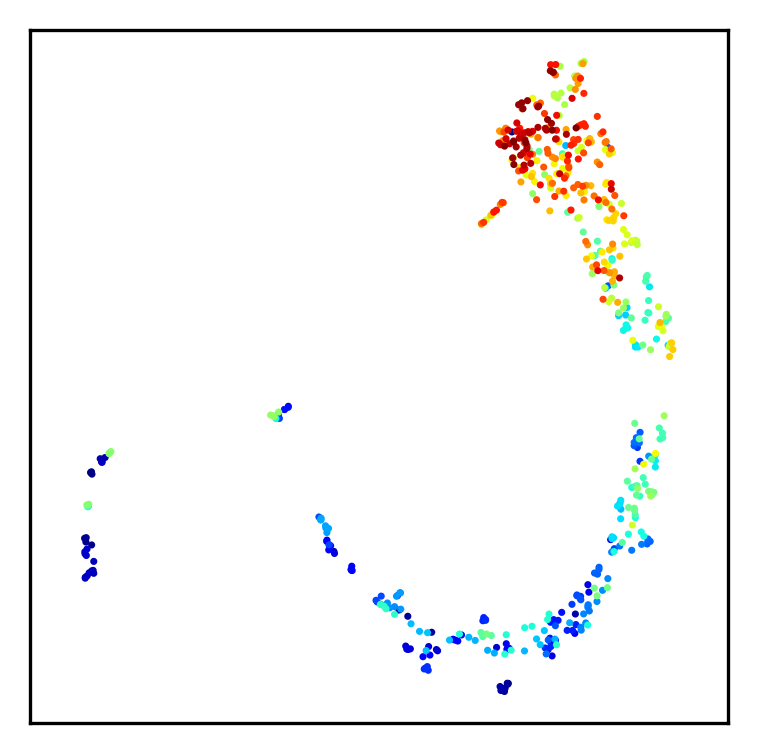}
        \caption{using RSS Distribution}
        \label{fig:subfig_a}
    \end{subfigure}
    \hfill
    \begin{subfigure}{0.23\textwidth}
    \centering
        \includegraphics[width=.8\textwidth]{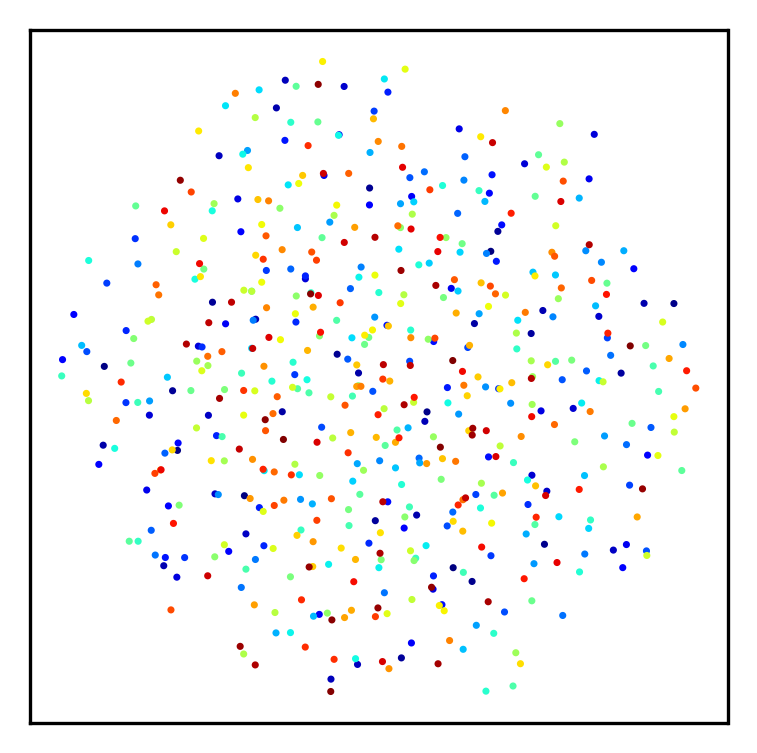}
        \caption{using Word Embedding\cite{mikolov2013efficient}}
        \label{fig:subfig_b}
    \end{subfigure}
    \caption{By using the RSS distribution for input embedding, the resultant MAC embeddings are more discriminative where we find physically closer MAC addresses (denoted in similar colors) are closer in the projected 2D feature space.}
    \label{fig:mapprnot}
\end{figure}

In our study conducted on Building A,  we compared the use of the RSS Distribution Map against directly using MAC address indices, akin to neural word embeddings. \cite{mikolov2013efficient}. The results revealed a notable increase in testing mean error by approximately 1.5 m, suggesting the model was overfitting to the radio map. The introduction of the RSS Distribution Map as an input appears to serve both as an initializer and as a regularizer. As shown in Fig.\ref{fig:mapprnot}, we present a 2D projection of the input embeddings using UMAP\cite{mcinnes2020umap} for ease of visualization. we demonstrated that MACs detected in close temporal proximity were similarly colored. The RSS Distribution Map enabled the model to capture both global and local topological structures and correlations between proximate MACs, unlike the baseline approach, which isolated MACs into separate spaces.

\subsubsection{Uncertainty Estimation}

We conducted another experiment on one building to examine the impact of uncertainty on our models' performances. Table \ref{tab:unc-res} highlights the enhancement of over 10\% for both WiFi and Fusion positioning results. By using the metric of Pearson product-moment correlation, we found that the correlation between errors and uncertainty in all environments exceeded 0.3, indicating a relatively strong relationship. Incorporating uncertainty offers dual advantages: it prevents the WiFi model from overfitting to the noisy radio map data and also refines the precision of Fusion's location predictions. The fusion model trusts the WiFi-based location estimates more while remaining cautious of those with high uncertainty.

\begin{table}
    \centering
    \caption{Uncertainty Integration Performance Comparison}
    \label{tab:unc-res}
    \begin{threeparttable}
    
\begin{tabular}{l|cccc}
\hline
Methods                    
                                           & Mean   & Median & Max     & 95th                  \\ \hline
WiFi (with Unc)                            & \textbf{2.5706} & \textbf{2.1238} & \textbf{14.4922} & \textbf{6.3661}               \\ 
WiFi (without Unc)                         & 3.1406 & 2.6301 & 21.9450 & 8.0191              \\ 
Fusion (with Unc)                          & \textbf{1.8558} &  \textbf{1.4207}    &    7.2425     & \textbf{4.1152}    \\ 
Fusion (without Unc)                       & 2.1558 &   1.5880  & \textbf{7.1923}        &    4.8716    \\ \hline
\end{tabular}
    \end{threeparttable}     
\end{table}

\section{Conclusion}
This research explored the potential of replacing traditional floor plans in Indoor Positioning Systems (IPS) with crowdsourced radio maps. Our findings highlighted radio maps as a viable alternative, given their inherent movement regularities. Despite challenges like inaccuracies and sparse coverage, our proposed framework, combining an uncertainty-aware neural network with Bayesian filtering, showcased a promising improvement of $\sim25\%$ over current IPS methods. As the digital landscape evolves, leveraging radio maps could redefine indoor localization. Future endeavors will focus on refining uncertainty models and exploring advanced fusion techniques to further enhance IPS efficacy.

\vspace{1em}
\bibliographystyle{IEEEtran}
\bibliography{MyLibrary}

\begin{thebibliography}{10}
\providecommand{\url}[1]{#1}
\csname url@rmstyle\endcsname
\providecommand{\newblock}{\relax}
\providecommand{\bibinfo}[2]{#2}
\providecommand\BIBentrySTDinterwordspacing{\spaceskip=0pt\relax}
\providecommand\BIBentryALTinterwordstretchfactor{4}
\providecommand\BIBentryALTinterwordspacing{\spaceskip=\fontdimen2\font plus
\BIBentryALTinterwordstretchfactor\fontdimen3\font minus
  \fontdimen4\font\relax}
\providecommand\BIBforeignlanguage[2]{{%
\expandafter\ifx\csname l@#1\endcsname\relax
\typeout{** WARNING: IEEEtran.bst: No hyphenation pattern has been}%
\typeout{** loaded for the language `#1'. Using the pattern for}%
\typeout{** the default language instead.}%
\else
\language=\csname l@#1\endcsname
\fi
#2}}

\bibitem{herath2021fusiondhl}
S.~Herath, S.~Irandoust, B.~Chen, Y.~Qian, P.~Kim, and Y.~Furukawa,
  ``Fusion-dhl: Wifi, imu, and floorplan fusion for dense history of locations
  in indoor environments,'' 2021.

\bibitem{xiao2014lightweight}
Z.~Xiao, H.~Wen, A.~Markham, and N.~Trigoni, ``Lightweight map matching for
  indoor localisation using conditional random fields,'' in \emph{IPSN-14
  proceedings of the 13th international symposium on information processing in
  sensor networks}.\hskip 1em plus 0.5em minus 0.4em\relax IEEE, 2014, pp.
  131--142.

\bibitem{huang2019online}
B.~Huang, Z.~Xu, B.~Jia, and G.~Mao, ``An online radio map update scheme for
  wifi fingerprint-based localization,'' \emph{IEEE Internet of Things
  Journal}, vol.~6, no.~4, pp. 6909--6918, 2019.

\bibitem{wifireview2015yang}
C.~Yang and H.-r. Shao, ``Wifi-based indoor positioning,'' \emph{IEEE
  Communications Magazine}, vol.~53, no.~3, pp. 150--157, 2015.

\bibitem{bluetoo2016chen}
Z.~Chen, Q.~Zhu, and Y.~C. Soh, ``Smartphone inertial sensor-based indoor
  localization and tracking with ibeacon corrections,'' \emph{IEEE Transactions
  on Industrial Informatics}, vol.~12, no.~4, pp. 1540--1549, 2016.

\bibitem{WiGLEonline}
\BIBentryALTinterwordspacing
Wigle - wireless geographic logging engine. [Online]. Available:
  \url{https://wigle.net/}
\BIBentrySTDinterwordspacing

\bibitem{radar2000bah}
P.~Bahl and V.~Padmanabhan, ``Radar: an in-building rf-based user location and
  tracking system,'' in \emph{Proceedings IEEE INFOCOM 2000. Conference on
  Computer Communications. Nineteenth Annual Joint Conference of the IEEE
  Computer and Communications Societies (Cat. No.00CH37064)}, vol.~2, 2000, pp.
  775--784 vol.2.

\bibitem{WILL2013wu}
C.~Wu, Z.~Yang, Y.~Liu, and W.~Xi, ``Will: Wireless indoor localization without
  site survey,'' \emph{IEEE Transactions on Parallel and Distributed Systems},
  vol.~24, no.~4, pp. 839--848, 2013.

\bibitem{crowdsource2010mink}
M.~Lee, H.~Yang, D.~Han, and C.~Yu, ``Crowdsourced radiomap for room-level
  place recognition in urban environment,'' \emph{2010 8th IEEE International
  Conference on Pervasive Computing and Communications Workshops (PERCOM
  Workshops)}, pp. 648--653, 2010.

\bibitem{kendall2017uncertainties}
A.~Kendall and Y.~Gal, ``What uncertainties do we need in bayesian deep
  learning for computer vision?'' 2017.

\bibitem{cai2022stun}
K.~Cai, C.~X. Lu, and X.~Huang, ``Stun: Self-teaching uncertainty estimation
  for place recognition,'' 2022.

\bibitem{poggi2020uncertainty}
M.~Poggi, F.~Aleotti, F.~Tosi, and S.~Mattoccia, ``On the uncertainty of
  self-supervised monocular depth estimation,'' 2020.

\bibitem{taha2019unsupervised}
A.~Taha, Y.-T. Chen, T.~Misu, A.~Shrivastava, and L.~Davis, ``Unsupervised data
  uncertainty learning in visual retrieval systems,'' 2019.

\bibitem{feng2018safe}
D.~Feng, L.~Rosenbaum, and K.~Dietmayer, ``Towards safe autonomous driving:
  Capture uncertainty in the deep neural network for lidar 3d vehicle
  detection,'' 2018.

\bibitem{ferris2007wifi}
B.~Ferris, D.~Fox, and N.~D. Lawrence, ``Wifi-slam using gaussian process
  latent variable models.'' in \emph{IJCAI}, vol.~7, no.~1, 2007, pp.
  2480--2485.

\bibitem{mirowski2013signalslam}
P.~Mirowski, T.~K. Ho, S.~Yi, and M.~MacDonald, ``Signalslam: Simultaneous
  localization and mapping with mixed wifi, bluetooth, lte and magnetic
  signals,'' in \emph{International Conference on Indoor Positioning and Indoor
  Navigation}.\hskip 1em plus 0.5em minus 0.4em\relax IEEE, 2013, pp. 1--10.

\bibitem{zou2017Accurate}
H.~Zou, Z.~Chen, H.~Jiang, L.~Xie, and C.~Spanos, ``Accurate indoor
  localization and tracking using mobile phone inertial sensors, wifi and
  ibeacon,'' in \emph{2017 IEEE International Symposium on Inertial Sensors and
  Systems (INERTIAL)}, 2017, pp. 1--4.

\bibitem{Hong2014WaP}
F.~Hong, Y.~Zhang, Z.~Zhang, M.~Wei, Y.~Feng, and Z.~Guo, ``Wap: Indoor
  localization and tracking using wifi-assisted particle filter,'' in
  \emph{39th Annual IEEE Conference on Local Computer Networks}, 2014, pp.
  210--217.

\bibitem{carrera2016real}
J.~L. Carrera, Z.~Zhao, T.~Braun, and Z.~Li, ``A real-time indoor tracking
  system by fusing inertial sensor, radio signal and floor plan,'' in
  \emph{2016 International Conference on Indoor Positioning and Indoor
  Navigation (IPIN)}.\hskip 1em plus 0.5em minus 0.4em\relax IEEE, 2016, pp.
  1--8.

\bibitem{Hellmers2013IMU}
H.~Hellmers, A.~Norrdine, J.~Blankenbach, and A.~Eichhorn, ``An
  imu/magnetometer-based indoor positioning system using kalman filtering,'' in
  \emph{International Conference on Indoor Positioning and Indoor Navigation},
  2013, pp. 1--9.

\bibitem{Poulose2019Indoor}
A.~Poulose and D.~S. Han, ``Indoor localization using pdr with wi-fi weighted
  path loss algorithm,'' in \emph{2019 International Conference on Information
  and Communication Technology Convergence (ICTC)}, 2019, pp. 689--693.

\bibitem{Feng2020Kalman}
D.~Feng, C.~Wang, C.~He, Y.~Zhuang, and X.-G. Xia, ``Kalman-filter-based
  integration of imu and uwb for high-accuracy indoor positioning and
  navigation,'' \emph{IEEE Internet of Things Journal}, vol.~7, no.~4, pp.
  3133--3146, 2020.

\bibitem{ma2017pedestrian}
L.~Ma, Y.~Fan, Y.~Xu, and Y.~Cui, ``Pedestrian dead reckoning trajectory
  matching method for radio map crowdsourcing building in wifi indoor
  positioning system,'' in \emph{2017 IEEE International Conference on
  Communications (ICC)}.\hskip 1em plus 0.5em minus 0.4em\relax IEEE, 2017, pp.
  1--6.

\bibitem{imupdr2013Li}
M.~Li and A.~Mourikis, ``High-precision, consistent ekf-based visual–inertial
  odometry,'' \emph{The International Journal of Robotics Research}, vol.~32,
  pp. 690--711, 05 2013.

\bibitem{wifioverview2017xia}
\BIBentryALTinterwordspacing
S.~Xia, Y.~Liu, G.~Yuan, M.~Zhu, and Z.~Wang, ``Indoor fingerprint positioning
  based on wi-fi: An overview,'' \emph{ISPRS International Journal of
  Geo-Information}, vol.~6, no.~5, 2017. [Online]. Available:
  \url{https://www.mdpi.com/2220-9964/6/5/135}
\BIBentrySTDinterwordspacing

\bibitem{vaswani2017attention}
A.~Vaswani, N.~Shazeer, N.~Parmar, J.~Uszkoreit, L.~Jones, A.~N. Gomez,
  {\L}.~Kaiser, and I.~Polosukhin, ``Attention is all you need,''
  \emph{Advances in neural information processing systems}, vol.~30, 2017.

\bibitem{devlin2018bert}
J.~Devlin, M.-W. Chang, K.~Lee, and K.~Toutanova, ``Bert: Pre-training of deep
  bidirectional transformers for language understanding,'' \emph{arXiv preprint
  arXiv:1810.04805}, 2018.

\bibitem{mikolov2013efficient}
T.~Mikolov, K.~Chen, G.~Corrado, and J.~Dean, ``Efficient estimation of word
  representations in vector space,'' 2013.

\bibitem{silverman2018density}
B.~W. Silverman, \emph{Density estimation for statistics and data
  analysis}.\hskip 1em plus 0.5em minus 0.4em\relax Routledge, 2018.

\bibitem{Thrun2005Probabilistic}
S.~Thrun, W.~Burgard, and D.~Fox, ``Probabilistic robotics,'' 2005.

\bibitem{hester2008negative}
T.~Hester and P.~Stone, ``Negative information and line observations for monte
  carlo localization,'' in \emph{2008 IEEE International Conference on Robotics
  and Automation}.\hskip 1em plus 0.5em minus 0.4em\relax IEEE, 2008, pp.
  2764--2769.

\bibitem{doucet2001sequential}
A.~Doucet, N.~De~Freitas, N.~J. Gordon, \emph{et~al.}, \emph{Sequential Monte
  Carlo methods in practice}.\hskip 1em plus 0.5em minus 0.4em\relax Springer,
  2001, vol.~1, no.~2.

\bibitem{huaweiradiomap}
I.~V. Rory~Hughes, Lei~Tao and F.~Alsehly, ``Calibration-free radiomap
  construction based on graph map matching,'' 2023.

\bibitem{liosam2020shan}
T.~Shan, B.~Englot, D.~Meyers, W.~Wang, C.~Ratti, and R.~Daniela, ``Lio-sam:
  Tightly-coupled lidar inertial odometry via smoothing and mapping,'' in
  \emph{IEEE/RSJ International Conference on Intelligent Robots and Systems
  (IROS)}.\hskip 1em plus 0.5em minus 0.4em\relax IEEE, 2020, pp. 5135--5142.

\bibitem{adamw2019decoupled}
I.~Loshchilov and F.~Hutter, ``Decoupled weight decay regularization,'' 2019.

\bibitem{relu2019deep}
A.~F. Agarap, ``Deep learning using rectified linear units (relu),'' 2019.

\bibitem{kalman1960new}
R.~E. Kalman, ``A new approach to linear filtering and prediction problems,''
  1960.

\bibitem{wknn2004he}
K.~Hechenbichler and K.~Schliep, ``Weighted k-nearest-neighbor techniques and
  ordinal classification,'' \emph{discussion paper}, vol. 399, 01 2004.

\bibitem{DSAE2019wa}
\BIBentryALTinterwordspacing
R.~Wang, Z.~Li, H.~Luo, F.~Zhao, W.~Shao, and Q.~Wang, ``A robust wi-fi
  fingerprint positioning algorithm using stacked denoising autoencoder and
  multi-layer perceptron,'' \emph{Remote Sensing}, vol.~11, no.~11, 2019.
  [Online]. Available: \url{https://www.mdpi.com/2072-4292/11/11/1293}
\BIBentrySTDinterwordspacing

\bibitem{mcinnes2020umap}
L.~McInnes, J.~Healy, and J.~Melville, ``Umap: Uniform manifold approximation
  and projection for dimension reduction,'' 2020.

\end{thebibliography}
\end{document}